\documentclass[pdflatex,sn-mathphys-ay]{sn-jnl}% Math and Physical Sciences Author Year Reference Style
\usepackage{graphicx}%
\usepackage{multirow}%
\usepackage{amsmath,amssymb,amsfonts}%
\usepackage{amsthm}%
\usepackage{mathrsfs}%
\usepackage[title]{appendix}%
\usepackage{textcomp}%
\usepackage{manyfoot}%
\usepackage{booktabs}%
\usepackage{algorithm}%
\usepackage{algorithmicx}%
\usepackage{algpseudocode}%
\usepackage{listings}%
\usepackage{bm}
\usepackage{pifont}
\usepackage{multirow}
\usepackage{subcaption}
\usepackage{threeparttable}
\usepackage{array}
\usepackage[table]{xcolor} % For coloring, should be removed

\usepackage{ulem} % For strike-through in revision
\usepackage{makecell}

\theoremstyle{thmstyleone}%
\theoremstyle{thmstyletwo}%

\theoremstyle{thmstylethree}%

\renewcommand{\Comment}[1]{\texttt{// #1}}

\begin{document}

\title[TRIP-Shapley: Propagation-Aware Contribution Attribution Framework for Peer-to-Peer Federated Learning]{TRIP-Shapley: Propagation-Aware Contribution Attribution Framework for Peer-to-Peer Federated Learning}

%%=============================================================%%
%% GivenName	-> \fnm{Joergen W.}
%% Particle	-> \spfx{van der} -> surname prefix
%% FamilyName	-> \sur{Ploeg}
%% Suffix	-> \sfx{IV}
%% \author*[1,2]{\fnm{Joergen W.} \spfx{van der} \sur{Ploeg} 
%%  \sfx{IV}}\email{iauthor@gmail.com}
%%=============================================================%%

\author*[1,2]{\fnm{Honoka} \sur{Anada}}\email{honokaanada@is.s.u-tokyo.ac.jp}

\author[3]{\fnm{Tatsuya} \sur{Kaneko}}\email{kaneko@artic.iir.isct.ac.jp}

\author[2]{\fnm{Yafei} \sur{Ou}}\email{yafei.ou@riken.jp}

\author[1,2]{\fnm{Shinya} \sur{Takamaeda-Yamazaki}}\email{shinya@is.s.u-tokyo.ac.jp}

\affil[1]{
\orgname{The University of Tokyo},
\orgaddress{\city{Tokyo},
\country{Japan}}}

\affil[2]{\orgname{RIKEN Center for Advanced Intelligence Project (AIP)},
\orgaddress{\city{Tokyo},
\country{Japan}}}

\affil[3]{
\orgname{Institute of Science Tokyo},
\orgaddress{\city{Yokohama},
\country{Japan}}}

\abstract{
Contribution attribution is a fundamental problem in federated learning (FL), yet it remains challenging in peer-to-peer FL (P2PFL), where participant influence propagates through repeated model exchanges. Recent work, DICE, addresses this issue by directly tracking multi-round propagation paths. However, its practical deployment is hindered by two challenges: (i) communication explosion caused by multi-hop propagation path tracking, and (ii) vulnerability to misreporting attacks. To overcome these challenges, we consider contribution attribution in P2PFL with a model-free coordinator used solely for auditing. Based on this setting, we propose TRIP-Shapley, a propagation-aware Shapley-inspired surrogate for contribution attribution. Instead of explicitly tracking multi-round paths, TRIP-Shapley computes round-wise contribution scores independently and propagates only scalar contribution values. This enables attribution of influence accumulated throughout the entire training process while maintaining practical communication costs. In addition, we introduce lightweight mechanisms to detect and mitigate misreporting attacks. Rather than theoretically approximating Shapley values, TRIP-Shapley is empirically evaluated against Shapley values as a reference. Experiments demonstrate communication savings of up to $45.8\times$ relative to the compared DICE configurations, with favorable cost--reference-alignment trade-offs overall, although alignment is weaker in some FMNIST and 25-client settings. The proposed defenses reduce contribution inflation under the two evaluated attacks.
}


%%\pacs[JEL Classification]{D8, H51}

%%\pacs[MSC Classification]{35A01, 65L10, 65L12, 65L20, 65L70}

\maketitle
\section{Introduction} \label{sec:introduction}

Federated learning (FL) \citep{fl-original} has emerged as an effective paradigm for privacy-preserving collaborative machine learning across distributed data sources without sharing raw data. While most existing studies focus on centralized FL (CFL) with a central parameter server \citep{fl-original}, recent advances in peer-to-peer FL (P2PFL) \citep{lalitha_fully_2018} have explored decentralized model exchange protocols that alleviate the communication bottleneck of a central aggregator.

A major challenge in FL is motivating data holders to actively participate in training. Since participation incurs computational costs and privacy risks, providing fair rewards is essential for sustaining participation \citep{liu_fedcoin_2020}, necessitating a contribution attribution method for reward assignment \citep{lv_data-free_2021}. Beyond incentive provision, contribution attribution is also important for improving the transparency and accountability of the collaborative training process.

Contribution attribution remains equally important in P2PFL, but becomes substantially more challenging. Unlike CFL, P2PFL lacks a parameter server that directly observes all model exchanges throughout the training process. Furthermore, a participant's influence is not confined to its immediate neighbors but propagates through repeated peer-to-peer exchanges across the communication graph. Consequently, accurately assessing contributions in P2PFL requires tracing how influence propagates through the network while relying only on information available to individual participants. However, most existing contribution attribution frameworks are targeted at CFL, failing to address these challenges.

Recent work has begun to address contribution attribution in this setting. For example, DICE \citep{zhu_dice_2025} formulates influence propagation as a multi-round Leave-One-Out (LOO) score and directly approximates it through a Hessian chain, with a configurable hop limit to capture influence over longer communication paths.

While DICE represents an important step toward propagation-aware contribution attribution in P2PFL, its practical deployment is hindered by two challenges. \textit{First}, DICE suffers from communication explosion due to multi-hop path tracking. Larger hop limits capture longer-range influence but exponentially increase tracked propagation paths and communication costs, while smaller limits reduce overhead at the cost of attribution accuracy. \textit{Second}, DICE is vulnerable to misreporting attacks. Its reliance on locally reported information allows malicious clients to distort attributed contributions through manipulated reports.

These challenges reveal a fundamental requirement for contribution attribution in P2PFL: a global view of influence propagation is required, though each client only observes local interactions. Achieving this in a fully decentralized setting would require Byzantine-tolerant consensus among all participants, whose communication and synchronization costs become impractical at scale. We therefore adopt a lightweight model-free coordinator as an auditing entity to enforce consistency and robustness without introducing a central parameter server.

\begin{figure}

\centering

\includegraphics[width=0.8\linewidth]{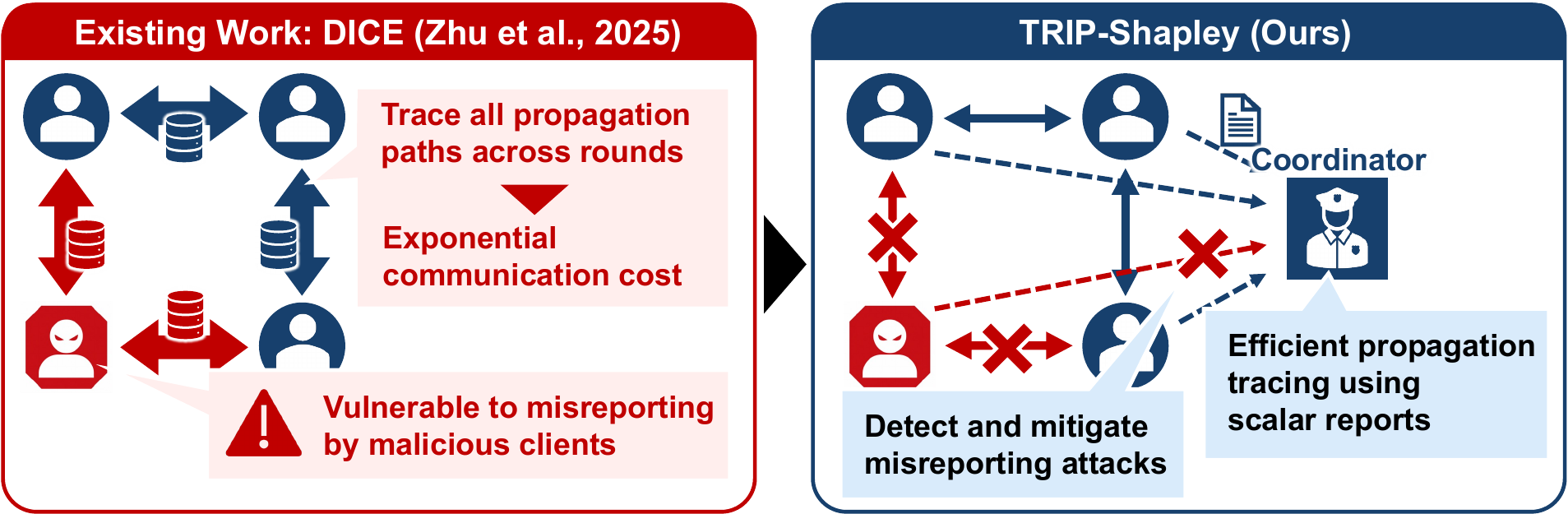}

\caption{While DICE suffers from two challenges, (i) communication explosion due to multi-hop path tracking and (ii) vulnerability to misreporting attacks, TRIP-Shapley addresses them by (i) tracing propagation using scalar contribution values and (ii) incorporating mechanisms to detect and mitigate misreporting attacks.}

\vspace{-8pt}
\label{fig:teaser}

\end{figure}

In this setting, we propose \textbf{TRIP-Shapley}, a propagation-aware, Shapley-inspired surrogate for contribution attribution for P2PFL, as shown in Fig.~\ref{fig:teaser}. The contributions of this work are summarized as follows:

\begin{itemize}

\item We propose TRIP-Shapley, which computes round-wise contribution scores independently and propagates scalar contribution values over the communication graph, capturing longer propagation paths than DICE while reducing communication by up to $45.8\times$.

\item We identify misreporting of contribution-related information as a practical vulnerability in decentralized contribution accounting. We analyze two representative attacks, namely LCV Misreporting (LCV Attack) and Pre-Training Model Falsification (PMF Attack), and introduce detection and mitigation mechanisms that improve robustness against these two attacks.

\item We introduce extensions of TRIP-Shapley that further reduce computational and communication overhead, whose effects on reference alignment are setting-dependent.

\item In 8-client experiments with varying topologies, TRIP-Shapley achieves favorable trade-offs between cost and reference rank alignment compared with DICE. However, alignment is weaker in some FMNIST and 25-client settings. The proposed defenses reduce contribution inflation under the evaluated attacks. Vanilla local attribution has an evaluation cost that is exponential in neighborhood size, whereas LOO reduces it to linear.

\end{itemize}

\section{Background and Related Work} \label{sec:background}

\subsection{Contribution Attribution in CFL}

As contribution attribution is essential for incentivizing active participation and enhancing transparency, numerous approaches have been proposed in prior studies. A major class of approaches leverages game-theoretic metrics such as the Shapley value \citep{shapley_value_1953, song_profit_2019, liu_gtg-shapley_2022}, Leave-One-Out \citep{wang_measure_2019, xue_toward_2021}, and LeastCore \citep{yan_if_2021}. Meanwhile, other methods estimate contributions based on dataset prototypes \citep{guo_contribution_2024}, analysis using a shared model \citep{wang_fast_2024}, or similarity between clients' model updates \citep{lv_data-free_2021}.

Among these, Shapley value-inspired methods have been widely adopted for contribution attribution. Originally introduced for profit allocation in cooperative games, the Shapley value \citep{shapley_value_1953} provides a mathematically principled and fair way to distribute rewards based on each participant's marginal contribution. While contribution attribution does not have a universally established ground truth, we adopt the Shapley value as a reference for assessing attributed contributions, following its established use in prior game-theoretic approaches.

In a cooperative game with $n$ participants denoted as $N := \{1, 2, \dots, n\}$, the Shapley value $\phi_i$ for client $i$, representing the client $i$'s contribution, is defined as follows:

\begin{equation}
\phi_i := \sum_{S \subseteq N \backslash \{i\}} \frac{u(S \cup \{i\}) - u(S)}{{|N|-1 \choose |S|}} \label{eq/shapley}
\end{equation}

Here, $u(S)$ represents the \textit{utility} of a participant subset $S \subseteq N$, i.e., the overall profit when the game is played only with the subset $S$. In the context of FL, $u(S)$ is typically defined as the final model accuracy obtained when the entire training process is performed using only the client subset $S$ \citep{song_profit_2019, liu_gtg-shapley_2022}.

However, computing the exact Shapley value requires rerunning the entire FL process for every client subset $S$, which is prohibitively expensive in practice. To address this, prior work has explored Shapley-inspired surrogate metrics without repeated training. A widely adopted approach is to sum round-wise Shapley values \citep{song_profit_2019, liu_gtg-shapley_2022, wei_efficient_2020, wang_measure_2019}. In this formulation, contributions are estimated by aggregating single-round updates under different coalitions $S$, rather than executing full multi-round training, thereby eliminating retraining overhead. Another line of work approximates Shapley values via pseudo-reconstruction \citep{song_profit_2019}, where the server simulates partial-client training post hoc by selectively applying historical updates from a given coalition, enabling efficient evaluation without explicit retraining. Our work also follows this Shapley-inspired surrogate-based line.

\subsection{Contribution Attribution in P2PFL} \label{sec:background-P2PFL}

In P2PFL, each participant updates its own local model and exchanges it with neighbors. These neighbors subsequently propagate the received models to their own neighbors, spreading the contribution generated by a participant's local training to non-adjacent nodes. Accurately quantifying participant-level contributions in P2PFL, therefore, requires precise modeling of this \textit{peer-to-peer influence propagation} process.

Some studies have explored contribution attribution protocols in parameter-server-free settings. FedCoin \citep{liu_fedcoin_2020} and FLChain \citep{bao_flchain_2019} focus on blockchain-powered FL systems. Meanwhile, CYCle \citep{tastan_cycle_2025} considers decentralized training in an all-to-all communication graph. However, these studies do not address situations where peer-to-peer influence propagation occurs.

Among existing studies, DICE \citep{zhu_dice_2025} incorporates peer-to-peer propagation into contribution attribution. The authors first introduce DICE-GT, a ground-truth Leave-One-Out (LOO) metric tailored to peer-to-peer settings, and subsequently propose a tractable approximation, DICE-E. DICE allows flexible configuration of the hop count, where larger hop counts capture influence propagated through longer communication paths.

However, the DICE paper only provides a mathematical formulation of influence propagation and does not specify how contribution scores can be assigned in a practical decentralized system under data locality constraints. Moreover, its formulation suggests several inherent challenges for practical deployment:

\textbf{(i) Communication explosion from multi-hop path tracking.} Since DICE aims to directly approximate the multi-round LOO score, it requires tracking and exchanging model-parameter-sized intermediate results for all propagation paths whose lengths are within a predefined hop limit. Accurate contribution attribution requires a sufficiently large hop limit; however, increasing the hop limit exponentially increases the number of intermediate results to be exchanged, resulting in prohibitive communication overhead. For example, in a regular graph with degree $d$, $h$-hop DICE incurs a communication cost that is $1 + \sum_{i=0}^{h-1} d^i$ times that of vanilla P2PFL.

\textbf{(ii) Vulnerability to misreporting attacks.} In the absence of a node with access to all models, DICE scores must be computed collaboratively by clients. Consequently, attributed contributions can be manipulated if malicious clients report fabricated values to other nodes, raising concerns about the reliability of the resulting contribution scores.

\begin{figure*}
    \centering
    \includegraphics[width=\linewidth]{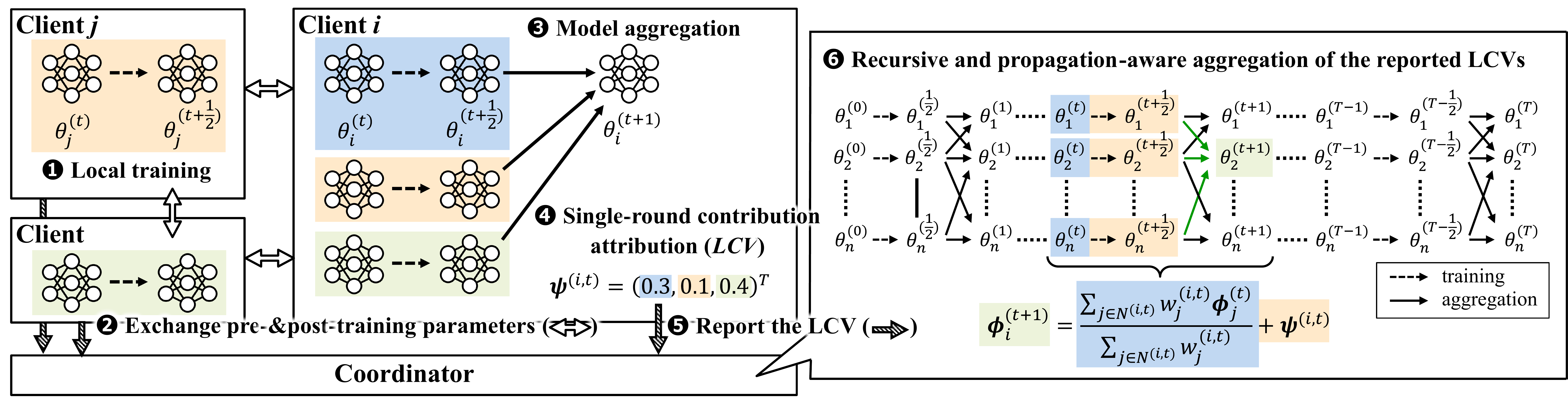}
    \caption{Overview of TRIP-Shapley: single-round contribution computation, LCV reporting, and propagation-aware aggregation.}
    \label{fig:trip}
    \vspace{-10pt}
\end{figure*}

\section{Methodology} \label{sec:proposal}

In this section, we introduce our Shapley-inspired surrogate contribution attribution framework, \underline{TR}acing \underline{I}nfluence \underline{P}ropagation Shapley (\textbf{TRIP-Shapley}). We first present the formulation of TRIP-Shapley under the assumption of honest participants, and then extend it with mechanisms to mitigate potential misreporting. Finally, we introduce further extensions for computational and communication cost reduction.

\subsection{Problem Formulation} \label{sec:formulation}

The peer-to-peer training protocol considered in this study encompasses a broad class of P2PFL algorithms, including those operating over asymmetric communication graphs \citep{parasnis_connectivity-aware_2023} and time-varying network topologies \citep{li_towards_2024, tang_gossipfl_2023}. There are $n$ clients, denoted by $N := \{1,2,\dots,n\}$. The P2PFL process consists of $T$ rounds in total, and all clients participate in every round. We denote the local model of client $i \in N$ at the beginning of round $t \in [0,T-1]$ as $\theta_i^{(t)}$. Each client possesses a private training dataset $\mathcal{D}_i$ and a shared test dataset $\mathcal{D}_\text{test}$, which is identical across all clients. In each round $t$, client $i$ trains its model $\theta_i^{(t)}$ on $\mathcal{D}_i$ to obtain $\theta_i^{(t+\frac{1}{2})}$, sends it to its out-neighbors $N_\text{out}^{(i,t)}$ of this round, receives $\theta_j^{(t+\frac{1}{2})}$ from its in-neighbors $N_\text{in}^{(i,t)}$, and aggregates its own and received models to obtain $\theta_i^{(t+1)}$.

We assume that model aggregation is performed by a weighted averaging of model parameters, as adopted in most previous studies \citep{fl-original, nedic_network_2018, tang_gossipfl_2023}, while other aggregation rules, including nonlinear or adaptive ones, are beyond the scope of this study. Specifically, the update rule for client $i$ at round $t$ is given by 
\begin{equation}
\theta_i^{(t+1)}
=
\sum_{j \in N^{(i,t)}} 
\frac{w_j^{(i,t)}}{\sum_{k \in N^{(i,t)}} w_k^{(i,t)}}
\, \theta_j^{\left(t+\frac{1}{2}\right)},
\label{eq:P2PFL-fedavg}
\end{equation}
where $N^{(i,t)} := N_{\mathrm{in}}^{(i,t)} \cup \{i\}$. Let $\Theta^{(t)} := \left(\theta_1^{(t)}, \theta_2^{(t)}, \dots, \theta_n^{(t)}\right)^\top$ denote the stacked vector of client models. We define the row-stochastic mixing matrix $W^{(t)} \in \mathbb{R}^{n \times n}$ whose $(i,j)$-th entry is given by:
\begin{equation}
W_{ij}^{(t)} :=
\begin{cases}
    % \displaystyle
    \frac{w_j^{(i,t)}}{\sum_{k \in N^{(i,t)}} w_k^{(i,t)}},
    & j \in N^{(i,t)}, \\[8pt]
    0, & \text{otherwise}.
\end{cases}
\end{equation}
Under this notation, the decentralized aggregation rule can be compactly written as:
\begin{equation} \label{eq:P2PFL-fedavg-compact}
\Theta^{(t+1)} = W^{(t)} \Theta^{(t+\frac{1}{2})} .
\end{equation}

In addition to normal nodes, a lightweight coordinator node is introduced solely for auditing purposes, as suggested in prior studies \citep{tang_gossipfl_2023, liao_adaptive_2023}. The coordinator monitors the P2PFL process with access \textit{only} to metadata regarding model exchange and aggregation, i.e., $N_\text{in}^{(i, t)}$, $N_\text{out}^{(i, t)}$, and $w_j^{(i,t)}$ for all clients, while never collecting model parameters. It also does not perform computationally intensive operations such as model training or evaluation. Consequently, the coordinator does not maintain a centralized model, while its additional contribution-accounting traffic is lightweight compared with model exchange. However, its aggregate communication and processing may become a scalability bottleneck as the number of clients grows.

Our objective is to have the coordinator derive a vector $\bm{\phi}_i^{(T)} \in \mathbb{R}^n$ for each client's final model $\theta_i^{(T)}$, where the $j$-th element of $\bm{\phi}_i^{(T)}$ quantifies client $j$'s contribution to $\theta_i^{(T)}$ throughout the entire P2PFL process.

We assume that the coordinator is always honest and that, for each client, a majority of its closed neighborhood, including itself, is honest. Since P2PFL contribution attribution relies on reciprocal neighbor-to-neighbor evaluation, settings violating this \textit{local honest-majority condition} are out of scope. We consider adversarial clients that may manipulate information used for contribution attribution. In contrast, protocol-level information, including client communication, topology, neighbor information, and aggregation weights, is assumed to be authenticated or verifiable. Clients are also assumed to follow the prescribed protocol for data reporting and reporting timing. Attacks involving falsification of protocol-level information, inconsistent reports to different peers, model poisoning, or Sybil attacks are also out of scope.

\subsection{TRIP-Shapley Design} \label{sec:proposal-honest}

First, we introduce the TRIP-Shapley design assuming all clients are honest. Figure~\ref{fig:trip} illustrates the procedure for contribution attribution in TRIP-Shapley. After local model training (\ding{"CA}), each client $i$ exchanges not only post-training models $\theta_i^{(t+\frac{1}{2})}$ but also pre-training models $\theta_i^{(t)}$, enabling the isolation of single-round contributions from those propagated across previous rounds (\ding{"CB}). Following model exchange and aggregation (\ding{"CC}), each client computes the single-round contributions $N^{(i,t)}$ of itself and its in-neighbors (\ding{"CD}). 

\textbf{Calculate LCVs (\ding{"CD}). }
We now elaborate on the definition of LCVs. For client $i$ and round $t$, LCV $\bm{\psi}^{(i,t)}$ is defined as the Shapley value of a local surrogate game over $N^{(i,t)}$, as formulated in Eq.~\ref{eq:lcv-definition}. Here, $u^{(i,t)}(S)$ is the utility function defined as the accuracy of the model $\theta_i^{(t+1)}(S)$ formulated as Eq.~\ref{eq:partial-model-for-lcv}.
\begin{equation}\label{eq:lcv-definition}
\bm{\psi}^{(i,t)}(j) :=
\begin{cases}
\sum_{S \subseteq N^{(i,t)} \backslash \{j\}} 
\frac{u^{(i,t)}(S \cup \{j\}) - u^{(i,t)}(S)}
{{|N^{(i,t)}|-1 \choose |S|}}, 
& j \in N^{(i,t)} \\
0, & \text{otherwise}
\end{cases}
\end{equation}
\begin{equation} \label{eq:partial-model-for-lcv}
\theta_i^{(t+1)}(S) := \frac{\sum_{j \in S} w_j^{(i,t)} \theta_j^{(t+\frac{1}{2})} + \sum_{j \in N^{(i,t)} \backslash S} w_j^{(i,t)} \theta_j^{(t)}}{\sum_{j \in N^{(i,t)}} w_j^{(i,t)}}
\end{equation}

This definition of LCVs mimics the round-wise Shapley value employed in prior studies~\citep{song_profit_2019}, as described in Section \ref{sec:background}. However, while existing methods define $u(S)$ as the accuracy of a model composed of post-training models for clients $i \in S$, our definition incorporates pre-training models from clients not in $S$. If we follow the same utility function definition as in CFL, the marginal contribution $u(S \cup \{i\}) - u(S)$ also includes indirect contributions from other clients that have been propagated from previous rounds. To isolate the effect of client $i$ in the current round, we must fix the contributions of other clients by replacing client $i$'s trained model with its pre-training model when it is not included in $S$.

These values are summarized as a \textit{Local Contribution Vector (LCV)} and reported to the coordinator (\ding{"CE}). The coordinator then performs propagation-aware aggregation over the reported LCVs to derive comprehensive contribution scores for all clients (\ding{"CF}). It should be noted that each LCV contains one scalar value per neighbor and is substantially smaller than the model parameters. However, collecting LCVs from all clients introduces additional communication that may become a scalability limitation as the system grows.

\textbf{Propagation-aware aggregation of LCVs (\ding{"CF}). }
We now describe how the coordinator estimates the overall contributions using the reported LCVs. Let $\bm{\phi}_i^{(t)}$ denote the cumulative contributions to $\theta_i^{(t)}$, i.e., to the local model of client $i$ at round $t$. The $j$-th element of $\bm{\phi}_i^{(t)}$ represents client $j$'s cumulative contribution to this model. Starting from $\bm{\phi}_i^{(0)} = \bm{0}$, the coordinator recursively applies the following update rule to obtain $\bm{\phi}_i^{(T)}$, representing the contributions over the entire training process:
\begin{equation}\label{eq:trip-update-rule}
   \bm{\phi}_i^{(t+1)} = \frac{\sum_{j \in N^{(i,t)}} w_j^{(i,t)} \bm{\phi}_j^{(t)}}{\sum_{j \in N^{(i,t)}} w_j^{(i,t)}} + \bm{\psi}^{(i,t)}
\end{equation}
Let $\Phi^{(t)} := (\bm{\phi}_1^{(t)}, \bm{\phi}_2^{(t)}, \dots, \bm{\phi}_n^{(t)})$ and $\Psi^{(t)} := (\bm{\psi}_1^{(t)}, \bm{\psi}_2^{(t)}, \dots, \bm{\psi}_n^{(t)})$ denote the stacked form of global contributions and LCVs, respectively. Using the mixing matrix $W^{(t)}$ defined in Section \ref{sec:background}, this update rule is compactly rewritten as:
\begin{equation}\label{eq:trip-update-rule-matrix}
    \Phi^{(t+1)} = W^{(t)} \Phi^{(t)} + \Psi^{(t)}
\end{equation}
\relax
The first term applies the same linear mixing process as model aggregation (Eq.~\ref{eq:P2PFL-fedavg-compact}), allowing past contribution information to propagate across the communication graph through peer-to-peer model exchanges. The second term injects newly generated local contributions at each round. Thus, the recursion algebraically follows the model-mixing dynamics, while the resulting contribution scores serve as a \textit{propagation-aware surrogate} for the final-model Shapley value rather than being guaranteed to equal it.

\subsection{Mitigation Against Misreporting Attacks} \label{sec:proposal-adversarial}

Next, we introduce mitigations to potential misreporting attacks by clients. Within the threat model in Section~\ref{sec:formulation}, we consider two concrete attacks: LCV misreporting (LCV Attack) and pre-training model falsification (PMF Attack). 
To detect and mitigate these attacks, we propose two corresponding defenses: pre-training model filtering and LCV outlier detection.

\textbf{LCV Attack: LCV Misreporting.}
Another possible attack is to send a fake LCV to the coordinator. An adversarial client might assign itself an artificially high value in its LCV to manipulate the coordinator into overestimating its contribution.

\textbf{Mitigation against LCV Attack: LCV Outlier Detection.}
To detect and correct misreported LCVs, we introduce an outlier detection mechanism. In each round, the coordinator examines all reported LCVs. If client $i$ misreports an LCV, for example, assigning itself an excessively high contribution, its LCV will deviate significantly from those of other clients, who do not report such a large value for client $i$. Our detection algorithm identifies such LCVs as outliers and adjusts them accordingly to maintain consistency across reports.

We now describe the detailed algorithm for outlier detection. Let $m$ be the number of non-zero elements across all reported LCVs, and let $\bm{p} \in \mathbb{R}^m$ collect these values, where the $k$-th element ($p_k$) is the contribution of client $j_k$ as reported by client $i_k$. Assuming that all honest clients report approximately consistent values for each client $j$ (\textit{cross-recipient LCV consistency assumption}), we let $v_j$ denote the \textit{true} contribution of client $j$. Thus, we expect $p_k \approx v_{j_k}$ for all $k$. This yields $\bm{p} = A\bm{v} + \bm{\gamma} + \bm{\epsilon}$; here, $A \in \mathbb{R}^{m \times n}$ maps $\bm{v}$ to the ideally computed LCVs $\hat{\bm{p}}$. Meanwhile, $\bm{\gamma} \in \mathbb{R}^m$ is a sparse vector encoding misreported values, where the $k$-th element is non-zero if and only if $p_k$ has been misreported. $\bm{\epsilon} \in \mathbb{R}^m$ represents small random noise.

We estimate $\bm{v}$ and misreported elements $\bm{\gamma}$ by solving
\begin{equation}
\min_{\bm{v} \in \mathbb{R}^n, \bm{\gamma} \in \mathbb{R}^m} \|\bm{p} - A\bm{v} - \bm{\gamma}\|_2^2 + \lambda \|\bm{\gamma}\|_1 \label{eq:lcv-outlier-target}
\end{equation}
The term $\lambda \|\bm{\gamma}\|_1$ penalizes non-sparsity in $\bm{\gamma}$, where $\lambda \in (0,1]$ controls its strength. Among various solvers proposed for this formulation of outlier detection, we adopt the $\Theta$-IPOD method \citep{she_outlier_2011} with soft thresholding, which iteratively optimizes $\bm{v}$ and $\bm{\gamma}$ alternately. 

After optimization, we construct a reference LCV without misreporting, denoted by $\hat{\bm{\psi}}^{(i,t)}$, whose $j$-th element is set to $v_j$ if $\bm{\psi}^{(i,t)}(j) \neq 0$, and to zero otherwise. Each reported LCV $\bm{\psi}^{(i,t)}$ is then compared against this reference vector; if the $\ell_2$ distance exceeds a predefined threshold $\tau$, it is flagged as misreported and replaced with $\hat{\bm{\psi}}^{(i,t)}$.

\textbf{PMF Attack: Pre-Training Model Falsification.}
In TRIP-Shapley, each client sends a model before training, $\theta_i^{(t)}$, which the recipient uses to isolate the client's marginal contribution in each round. An adversarial client may intentionally send a fake pre-training model with abnormally low accuracy, $\hat{\theta}_i^{(t)}$, thereby misleading the recipient into assigning it an inflated contribution value in the LCV.

\textbf{Mitigation against PMF Attack: Pre-Training Model Filtering.}
To prevent contribution manipulation through pre-training model falsification, we introduce pre-training model filtering. When an honest client $i$ receives pre-training models $\theta_j^{(t)}$ from all in-neighbors, it evaluates their accuracies, denoted $v_j^{(t)}$, as well as the accuracy of its own pre-training model, $v^{(t)}$. If $v_j^{(t)}$ is lower than $v^{(t)} - v_\text{threshold}$ for some fixed threshold $v_\text{threshold}$, the client regards $\theta_j^{(t)}$ as a falsified pre-training model with abnormally low accuracy and replaces it with its own pre-training model $\theta_i^{(t)}$ when calculating the LCV.

\begin{algorithm}[!t]
\caption{Overall procedure of TRIP-Shapley.}
\label{alg:proposal}
\begin{algorithmic}[1]

\Statex \textit{Client $i$:}
\State $\theta_i^{(0)} \gets$ Initial model parameters

\For{$t = 0$ to $T-1$}
\State $\theta_i^{(t+\frac{1}{2})} \gets \textsc{Train}(\theta_i^{(t)}, \mathcal{D}_i)$ \Comment{\ding{"CA}: Local training}
\State Exchange $\theta_i^{(t+\frac{1}{2})},\theta_i^{(t)}$
with $j\in N_\text{out}^{(i,t)}$ \Comment{\ding{"CB}: Model exchange}

\State $\theta_i^{(t+1)} \gets \left( \sum_{j\in N^{(i,t)}} w_j^{(i,t)} \theta_j^{(t+\frac{1}{2})} \right) \Big/ \left( \sum_{j\in N^{(i,t)}} w_j^{(i,t)} \right)$ \Comment{\ding{"CC}: Model aggr.}

\State \Comment{\ding{"CD}: Calculate an LCV}
\State $v^{(t)} \gets \textsc{Eval}(\theta_i^{(t)},\mathcal{D}_{\text{test}})$
\For{each $j\in N_\text{in}^{(i,t)}$}
    \If{$\textsc{Eval}(\theta_j^{(t)},\mathcal{D}_{\text{test}}) < v^{(t)} -v_{\text{threshold}}$}
        \State $\theta_j^{(t)} \gets \theta_i^{(t)}$
    \EndIf
\EndFor

\For{each $S\subseteq N^{(i,t)}$}
    \State Calculate $\theta_i^{(t+1)}(S)$ based on Eq.~\eqref{eq:partial-model-for-lcv}
    \State $u^{(i,t)}(S) \gets \textsc{Eval}(\theta_i^{(t+1)}(S),\mathcal{D}_{\text{test}})$
\EndFor

\State Compute $\bm{\psi}^{(i,t)}$ based on Eq.~\eqref{eq:lcv-definition}

\State Send $\bm{\psi}^{(i,t)}$ to coordinator \Comment{\ding{"CE}: Report LCV}

\EndFor

\Statex \textit{Coordinator:}
\State $\bm{\phi}_i^{(0)} \gets \bm{0}$ for each $i\in N$

\For{$t = 0$ to $T-1$}

\State Receive $\bm{\psi}^{(i,t)}$ from each $i\in N$

\State \Comment{\ding{"CF}: Aggregate the reported LCVs}
\State Run LCV outlier detection on $\{\bm{\psi}^{(i,t)}\mid i\in N\}$
\For{each $i\in N$}
    \State $\bm{\phi}_i^{(t+1)} \gets$ Eq.~\eqref{eq:trip-update-rule}
\EndFor

\EndFor

\end{algorithmic}
\end{algorithm}

\subsection{Procedure for Computing TRIP-Shapley} \label{sec:proposal-summary}

By combining the basic contribution attribution procedure defined in Section \ref{sec:proposal-honest} with the countermeasures against misreporting attacks introduced in Section \ref{sec:proposal-adversarial}, we arrive at the complete TRIP-Shapley protocol, as described in Algorithm \ref{alg:proposal}.

\subsection{Extension for Computational and Communication Cost Reduction} \label{sec:cost-reduction}

Despite its effectiveness in contribution attribution, vanilla TRIP-Shapley can introduce high computational and communication costs in large-scale settings. In this part, we analyze its overhead and introduce approximations and quantized model exchange to reduce it, with their effects on reference alignment evaluated empirically.

\textbf{Computational Cost. }
TRIP-Shapley requires each client to (i) compute LCVs by evaluating models for all neighbor coalitions and (ii) evaluate all received pre-training models to detect falsified ones. Hence, each client performs $d + 2^{d+1}$ model evaluations per round, which is substantial when the number of in-neighbors $d$ is large. This cost can be reduced using Monte Carlo-based Shapley approximations \citep{castro_polynomial_2009, jia_towards_2019}, with the simplest Leave-One-Out (LOO) approximation requiring only $2d + 1$ evaluations.

\textbf{Communication Cost. }
In vanilla TRIP-Shapley, clients exchange both pre- and post-training model parameters, doubling the communication volume compared to a naive implementation. This overhead can be reduced using gradient compression techniques for FL \citep{shlezinger_uveqfed_2021}, such as exchanging a post-training model along with a quantized difference between the post- and pre-training models.

\section{Experiment and Discussion} \label{sec:evaluation}

In this part, we conduct four groups of experiments to answer the following research questions:

\begin{itemize}

\item \textbf{RQ1}: Can TRIP-Shapley alleviate the trade-off between cost and reference alignment in existing approaches as explained in Section \ref{sec:background-P2PFL}?

\item \textbf{RQ2}: Does TRIP-Shapley achieve rank alignment with the reference Shapley value comparable to existing CFL methods?

\item \textbf{RQ3}: Is TRIP-Shapley robust against adversarial misreporting attacks?

\item \textbf{RQ4}: How do the rank alignment with the reference Shapley value and computational/communication costs of TRIP-Shapley scale as the system size increases?

\end{itemize}

We implement a P2PFL simulation following the procedure described in Section~\ref{sec:formulation}. We perform image classification tasks on CIFAR-10 \citep{cifar10} and FashionMNIST (FMNIST) \citep{fmnist}, using a lightweight convolutional neural network (CNN) model. For each dataset, the training set is divided into $n$ partitions, each serving as the local training dataset $\mathcal{D}_i$ for an individual client. We randomly select 500 entries from the test set to form a test dataset, $\mathcal{D}_\text{test}$. The cost–rank-alignment experiments in Section~\ref{sec:exp-comp-comm-tradeoff} (Fig.~\ref{fig:cost-tradeoff}) were repeated five times, and their means and standard deviations are reported. All other experiments were conducted once. Regarding defense, we use $v_\text{threshold}=0$, $\lambda=0.01$, and $\tau=0.005$ for all experiments except for sensitivity analysis in Section~\ref{sec:exp-dishonest}, with values chosen based on informal preliminary experiments and practical judgment rather than a predefined selection criterion.

\subsection{Rank Alignment with the Reference Shapley Value} \label{sec:exp-fidelity}

To answer RQ1 and RQ2, we evaluate the alignment of attributed contribution rankings with the reference Shapley value. Specifically, we measure Spearman's rank correlation coefficient between the attributed contribution scores and the reference Shapley values. In the reference Shapley value, for each coalition $S$, clients outside $S$ are replaced with dummy clients that perform no local training but continue model exchange, and the entire training process is rerun to obtain each client's final model accuracy as the utility. We adopt this definition because excluding clients entirely would alter the P2PFL communication graph and confound the effect of topology on contribution. Since its exact computation incurs exponential computational complexity, we restrict the experiments to 8 clients and 10 communication rounds.

\subsubsection{Trade-off between Cost and Reference Alignment Across Different Network Topologies} \label{sec:exp-comp-comm-tradeoff}

First, we investigate whether TRIP-Shapley alleviates the trade-off between cost and reference alignment inherent in existing approaches. Since the limitation of DICE primarily stems from its inability to fully capture influence propagation in asymmetric communication graphs, we vary the network topology while fixing the data distribution to IID to see if TRIP-Shapley performs better in capturing topological bias. We evaluate the following three communication topologies: (i) a star topology, in which a single hub client is connected to all other clients (\texttt{star}), (ii) a single-line topology (\texttt{line}), and (iii) two fully connected clusters connected by a single path (\texttt{cluster}). 

In addition to rank correlation with the reference Shapley value, we derive per-client and per-round communication volume and computational cost \textit{relative to those of vanilla P2PFL} without contribution attribution. Communication volume is measured as the per-round data volume sent by a single client to its peers and the coordinator, including original model exchanges, pre-training models/LCVs (for TRIP-Shapley), and intermediate vectors (for DICE). Computational cost is measured as the number of multiply-accumulate operations per client per round required for training and contribution computation, including model evaluation for LCV calculation and, for DICE, Hessian-vector-product computation.

We compare the baseline DICE, vanilla TRIP-Shapley, and TRIP-Shapley with communication and computational cost reduction extensions: 8-bit and 4-bit quantizations for communication reduction, and Monte-Carlo (MC) and LOO approximations for computation reduction. All methods share the same model size, topology, and communication accounting assumptions.

\begin{figure}[!t]
    \centering
    \begin{subfigure}[t]{0.23\linewidth}
        \centering
        \includegraphics[width=\linewidth]{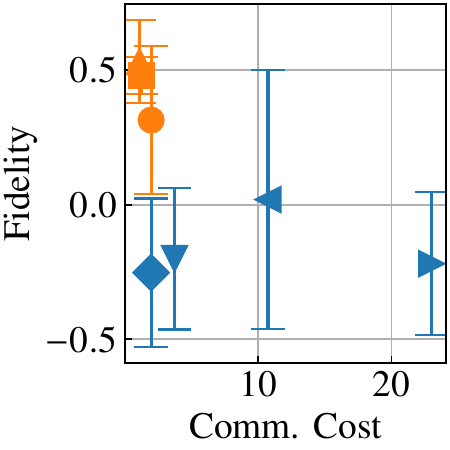}
        % \caption{star}
    \end{subfigure}
    \begin{subfigure}[t]{0.23\linewidth}
        \centering
        \includegraphics[width=\linewidth]{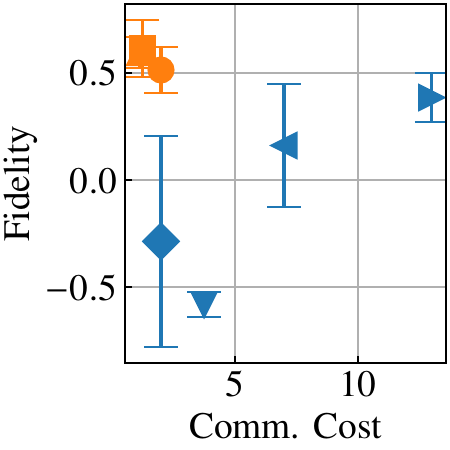}
        % \caption{line}
    \end{subfigure}
    \begin{subfigure}[t]{0.23\linewidth}
        \centering
        \includegraphics[width=\linewidth]{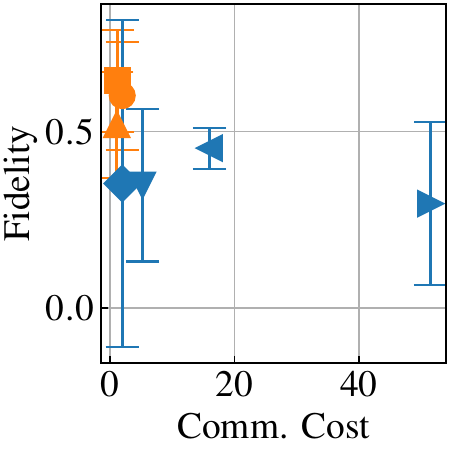}
        % \caption{cluster}
    \end{subfigure}
    \begin{subfigure}[t]{0.19\linewidth}
        \centering
        \includegraphics[width=\linewidth]{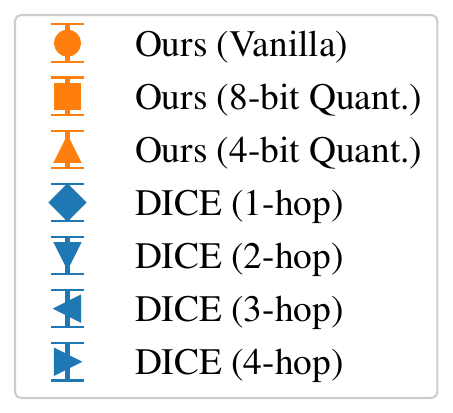}
    \end{subfigure}

    \vspace{-2pt}

    \begin{subfigure}[t]{0.23\linewidth}
        \centering
        \includegraphics[width=\linewidth]{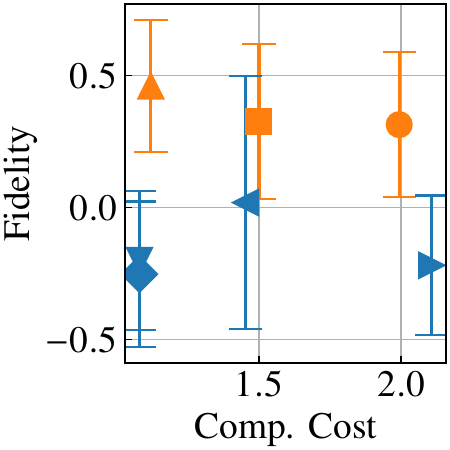}
        \caption{star}
    \end{subfigure}
    \begin{subfigure}[t]{0.23\linewidth}
        \centering
        \includegraphics[width=\linewidth]{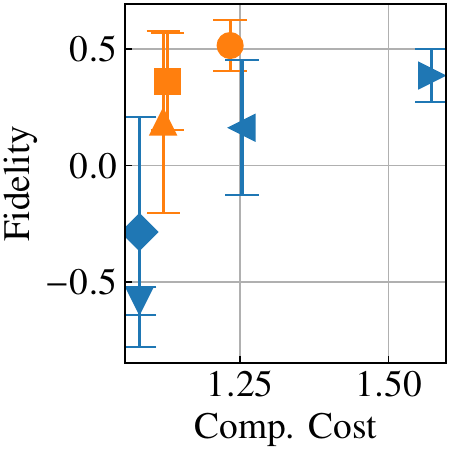}
        \caption{line}
    \end{subfigure}
    \begin{subfigure}[t]{0.23\linewidth}
        \centering
        \includegraphics[width=\linewidth]{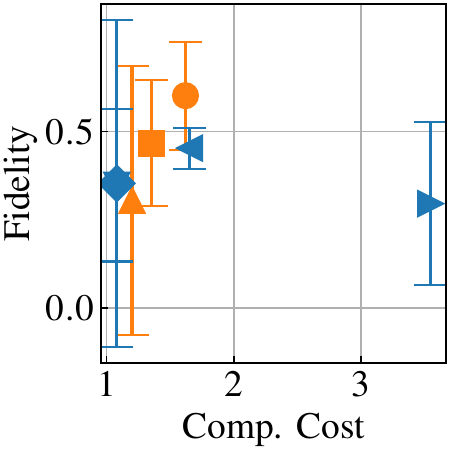}
        \caption{cluster}
    \end{subfigure}
    \begin{subfigure}[t]{0.19\linewidth}
        \centering
        \includegraphics[width=\linewidth]{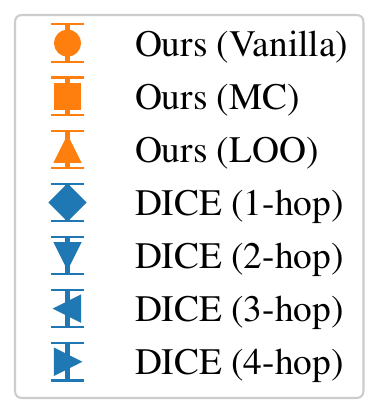}
    \end{subfigure}

    \caption{Communication (top) and computation (bottom) costs relative to vanilla P2PFL versus rank correlation with the reference Shapley value, measured by Spearman's rank correlation, for star, line, and cluster topologies.}
    
    \label{fig:cost-tradeoff}
    \vspace{-8pt}
\end{figure}

Figure~\ref{fig:cost-tradeoff} presents the trade-off between cost and reference alignment of the compared methods. For the DICE baseline, the one-hop variants exhibit near-zero or even negative Spearman's rank correlations, indicating little correspondence between the attributed scores and the reference Shapley values. In the \texttt{line} and \texttt{cluster} topologies, the correlation improves as the hop limit increases; however, this improvement comes at a substantial cost, particularly in terms of communication overhead, which reaches up to $51.5\times$ that of vanilla P2PFL in the four-hop setting. In contrast, TRIP-Shapley consistently maintains low communication overhead, requiring at most $2\times$ that of vanilla P2PFL, and as little as $1.13\times$ when 4-bit quantization is applied. Furthermore, Fig.~\ref{fig:cost-tradeoff} shows that TRIP-Shapley achieves higher rank correlation with the reference Shapley value than DICE variants with comparable computational costs. These results suggest that TRIP-Shapley effectively mitigates the severe trade-off between cost and reference alignment observed in DICE.

\subsubsection{Comparison with CFL Methods Across Different Data Distributions}

Next, we compare the rank alignment with the reference Shapley value with existing CFL methods under varying data distributions. The experiments are performed under five data distributions, following prior studies in CFL \citep{song_profit_2019, liu_gtg-shapley_2022}, while fixing the communication topology to a 4-regular graph: (i) IID and equal-sized datasets (\texttt{iid}); (ii) equal-sized but different label distributions (\texttt{non-iid}); (iii) IID but varying dataset sizes (\texttt{sizes}); (iv) images corrupted with various noise intensities (\texttt{noisy-images}); and (v) labels flipped with varying ratios (\texttt{noisy-labels}).

Table~\ref{tab:vs-shapley} shows the results. In CIFAR-10, TRIP-Shapley consistently achieves Spearman rank correlations above $0.5$, indicating positive alignment with the reference Shapley value. Meanwhile, with FMNIST, TRIP-Shapley shows lower correlation scores than DICE in \texttt{iid} and \texttt{noisy-labels} settings. In these two settings, the low cross-client contribution variance makes Spearman's correlations highly sensitive to minor score perturbations. As a complementary measure, TRIP-Shapley yields cosine distances of $0.007$ and $0.014$ in the two settings, compared with $0.009$/$0.308$ for one-hop DICE and $0.003$/$0.153$ for three-hop DICE. While these results indicate similarity in vector direction, they do not compensate for the low rank correlations and thus remain a limitation.

\begin{table}[!t]
    \centering
    \setlength{\tabcolsep}{2pt}
    \renewcommand{\arraystretch}{0.95}
    
    \caption{Spearman's rank correlations between TRIP-Shapley and the reference Shapley value are compared with existing methods for CFL (MR~\citep{song_profit_2019}, TMR~\citep{wei_efficient_2020}, GTG-Shapley~\citep{liu_gtg-shapley_2022}, and OR~\citep{song_profit_2019}) and for P2PFL (DICE~\citep{zhu_dice_2025}) across various data distributions. The best results in each column are highlighted with \underline{underline}.}
    
    \begin{tabular}{cclcccp{10mm}p{10mm}cccp{10mm}p{10mm}}
        \toprule
        &&& \multicolumn{5}{c}{CIFAR-10 $\uparrow$} & \multicolumn{5}{c}{FMNIST $\uparrow$} \\ 
        \cmidrule(lr){4-8} \cmidrule(lr){9-13}
        &&& iid & non-iid & sizes & noisy-images & noisy-labels & iid & non-iid & sizes & noisy-images & noisy-labels \\
        \midrule
        \multirow{4}{*}{\rotatebox[origin=c]{90}{CFL}} 
        & \multicolumn{2}{l}{MR}        
            & \underline{0.667} & 0.619 & 0.952 & 0.857 & -0.500 
            & 0.429 & 0.333 & 0.786 & 0.762 & -0.048 \\
        & \multicolumn{2}{l}{TMR}     
            & 0.524 & 0.405 & 0.952 & \underline{0.952} & \underline{0.595} 
            & -0.333 & 0.452 & 0.833 & -0.238 & 0.119 \\
        & \multicolumn{2}{l}{GTG-Shapley}   
            & -0.048 & -0.381 & -0.190 & 0.357 & 0.238 
            & \underline{0.833} & 0.619 & 0.190 & 0.071 & -0.190 \\
        & \multicolumn{2}{l}{OR}        
            & 0.000 & \underline{0.738} & \underline{0.976} & 0.619 & 0.429 
            & -0.024 & \underline{0.643} & \underline{0.905} & \underline{0.833} & \underline{0.786} \\
        \midrule
        \multirow{3}{*}{\rotatebox[origin=c]{90}{P2PFL}} 
        & \multicolumn{2}{l}{DICE (1-hop)}   
            & 0.119 & 0.405 & 0.952 & \underline{0.976} & \underline{0.881} 
            & \underline{0.429} & 0.524 & 0.881 & 0.143 & 0.929 \\
        & \multicolumn{2}{l}{DICE (3-hop)}   
            & -0.238 & -0.381 & \underline{0.976} & 0.905 & 0.810 
            & 0.119 & \underline{0.690} & 0.833 & \underline{0.476} & \underline{0.976} \\
        & \multicolumn{2}{l}{Ours}   
            & \underline{0.714} & \underline{0.810} & 0.952 & \underline{0.976} & 0.524 
            & 0.095 & 0.643 & \underline{0.976} & 0.405 & 0.000\\
        \bottomrule
    \end{tabular}
    \vspace{-12pt}
    \label{tab:vs-shapley}
\end{table}

\subsection{Robustness of TRIP-Shapley against Misreporting} \label{sec:exp-dishonest}

To answer RQ3, we evaluate TRIP-Shapley in the presence of misreporting clients with 25 clients. We randomly select $n_a$ clients to perform misreporting attacks. Adversarial clients are selected to fulfill the local honest-majority condition stated in Section \ref{sec:formulation}. These clients perform LCV and PMF attacks as defined in Section~\ref{sec:proposal-adversarial}, aiming to have the coordinator overestimate their contributions. In particular, we conduct experiments involving the following two types of attacks:

\begin{enumerate}
    \item[(i)] \textbf{Individual attack}: Each adversarial client attempts to increase only its own contribution score. Specifically, it adds a constant value $\alpha_a = 0.5$ to its own contribution score in the LCV for each round.
    
    \item[(ii)] \textbf{Coordinated attack}: Adversarial clients cooperate to inflate each other's contribution scores. Specifically, they add $\alpha_a = 0.5$ to all neighboring adversarial clients' scores in their LCVs.
\end{enumerate}

Regarding the PMF attack, adversarial clients always perform the individual attack, where they add Gaussian noise with standard deviation $\sigma_a = 0.05$ to their pre-training model parameters to degrade its performance. Meanwhile, more adaptive attackers that explicitly optimize attacks to evade detection are left for future work.

To quantitatively assess the extent of success of an attack under each condition, we introduce a metric named \textit{Contribution Inflation Factor} (CIF). Denoting the set of adversarial clients as $S_a$, and each client $i$'s contribution with and without manipulation as $\phi_i^*$ and $\phi_i$, the CIF $p_\text{CIF}$ for this manipulation attack is defined as $p_\text{CIF} = \sum_{i \in S_a} \phi_i^* / \sum_{i \in S_a} \phi_i$. Under this definition, larger CIF values indicate more successful manipulation by adversarial clients, whereas CIF equals 1 when the attack has no effect. We conduct experiments on the CIFAR-10 dataset with IID and Non-IID (Dirichlet with $\alpha=0.1$) data under varying numbers of adversarial clients. In addition to TRIP-Shapley with and without misreporting mitigation, we include one-hop DICE as a baseline. Both LCV and PMF attacks are also applicable to DICE, which can be implemented most naturally by having clients exchange their pre- and post-training models and evaluate their neighbors' contribution scores.

\begin{figure}[!t]
    \centering
    \begin{subfigure}{0.8\linewidth}
        \centering
        \includegraphics[width=\linewidth]{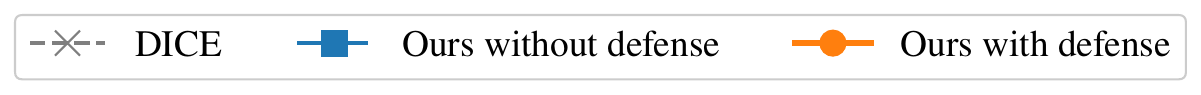}
    \end{subfigure}
    \begin{subfigure}{0.24\linewidth}
        \centering
        \includegraphics[width=\linewidth]{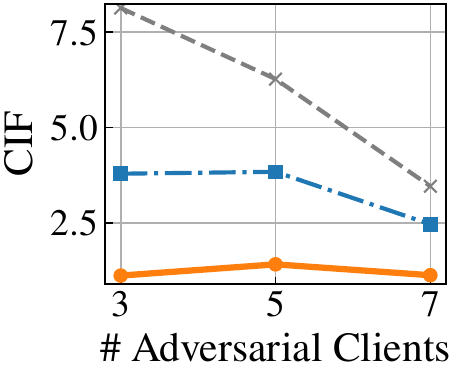}
        \caption{Indiv., IID}
        \label{fig:adversary-individual-iid}
    \end{subfigure}
    \begin{subfigure}{0.24\linewidth}
        \centering
        \includegraphics[width=\linewidth]{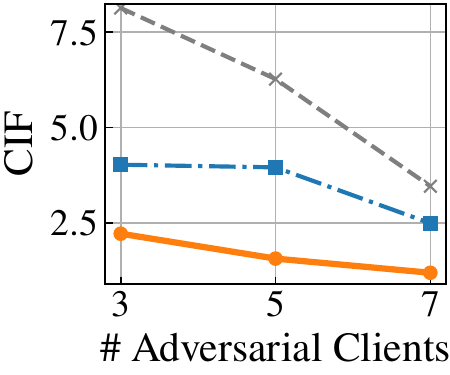}
        \caption{Coord., IID}
        \label{fig:adversary-coordinated-iid}
    \end{subfigure}
    \begin{subfigure}{0.24\linewidth}
        \centering
        \includegraphics[width=\linewidth]{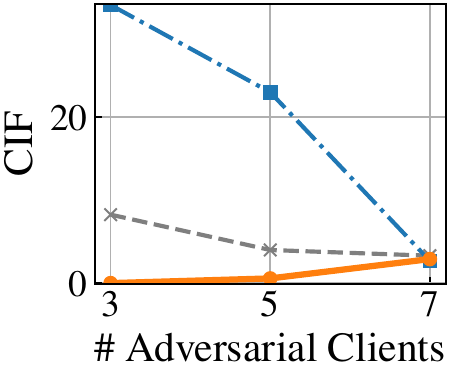}
        \caption{Indiv., Non-IID}
        \label{fig:adversary-individual-noniid}
    \end{subfigure}
    \begin{subfigure}{0.24\linewidth}
        \centering
        \includegraphics[width=\linewidth]{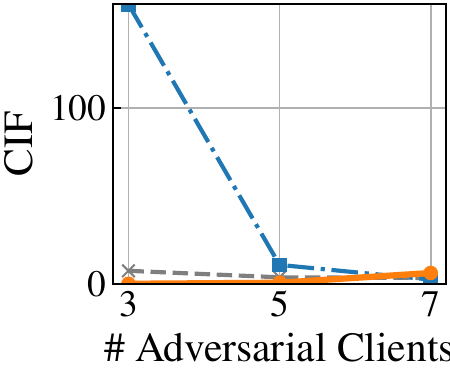}
        \caption{Coord., Non-IID}
        \label{fig:adversary-coordinated-noniid}
    \end{subfigure}

    \caption{Contribution Inflation Factors (CIFs) under individual and coordinated attacks with varying numbers of adversarial clients, where larger CIF values indicate more successful manipulation. TRIP-Shapley with mitigation consistently maintains lower CIF values than without mitigation or DICE.}
    \vspace{-10pt}
    
    \label{fig:adversary}
\end{figure}

\begin{table}[t]
    \centering
    \footnotesize
    \setlength{\tabcolsep}{1pt}
    \caption{CIF under different hyperparameter settings. Underlined values indicate the lowest CIF, while gray columns indicate the default settings.}
    \label{tab:adversary-defense-sensitivity}
    \begin{tabular}{
        ll|
        c
        >{\columncolor{gray!15}}c
        cc|
        c
        >{\columncolor{gray!15}}c
        cc|
        >{\columncolor{gray!15}}c
        cc||
        cc
        >{\columncolor{gray!15}}c
        c|
        cc
        >{\columncolor{gray!15}}c
        c
    }
        \toprule
        && \multicolumn{11}{c||}{Defense hyperparameters}
        & \multicolumn{8}{c}{Attack hyperparameters} \\
        \multirow{2}{*}{Data} & \multirow{2}{*}{Attack}
        & \multicolumn{4}{c|}{$\tau$}
        & \multicolumn{4}{c|}{$\lambda$}
        & \multicolumn{3}{c||}{$v_\mathrm{threshold}$}
        & \multicolumn{4}{c|}{$\alpha_a$}
        & \multicolumn{4}{c}{$\sigma_a$} \\
        & & .001 & .005 & .01 & .05
          & .005 & .01 & .05 & .1
          & 0 & .05 & .1
          & .1 & .3 & .5 & .7
          & .01 & .03 & .05 & .07 \\
        \midrule
        \multirow{2}{*}{IID} 
        & Indiv.
        & 1.51 & 1.41 & \underline{1.41} & 1.51
        & \underline{1.40} & 1.41 & 2.01 & 2.46
        & \underline{1.41} & 2.11 & 1.82
        & \underline{1.41} & \underline{1.41} & \underline{1.41} & \underline{1.41}
        & 1.15 & \underline{1.13} & 1.41 & 1.44 \\
        & Coord.
        & 1.61 & 1.57 & \underline{1.45} & 1.65
        & 2.94 & \underline{1.57} & 2.30 & 2.87
        & \underline{1.57} & 2.29 & 1.99
        & \underline{1.53} & 1.55 & 1.57 & 2.77
        & 1.32 & \underline{1.30} & 1.57 & 1.59 \\
        Non-
        & Indiv.
        & \underline{0.59} & 0.59 & 0.59 & 0.59
        & \underline{0.59} & 0.59 & 0.65 & 0.64
        & 0.59 & 0.38 & \underline{0.23}
        & \underline{0.59} & \underline{0.59} & \underline{0.59} & \underline{0.59}
        & 0.88 & 0.63 & \underline{0.59} & 0.66 \\
        IID
        & Coord.
        & 0.61 & 0.61 & \underline{0.60} & \underline{0.60}
        & 4.84 & \underline{0.61} & 0.78 & 1.24
        & 0.61 & 0.39 & \underline{0.22}
        & \underline{0.61} & 0.61 & 0.61 & 1.56
        & 0.86 & \underline{0.59} & 0.61 & 0.65 \\
        \bottomrule
    \end{tabular}
    \vspace{-7pt}
\end{table}

Figure~\ref{fig:adversary} shows the results. Under IID settings, DICE and TRIP-Shapley without mitigation exhibit CIF values as high as $8.13$ and $4.02$, respectively. In contrast, the proposed mitigations reduce the CIF by up to $3.39\times$ and keep it below $2.22$ across all settings. Under Non-IID settings, the CIF values without mitigation become substantially larger, reaching as high as $158.9$. Nevertheless, the proposed mitigations remain effective and drastically reduce the CIF. 

Additionally, we evaluate LCV outlier detection separately from the downstream CIF metric. No false positives or false negatives were observed under the evaluated individual attacks. Under coordinated attacks with 3, 5, and 7 adversarial clients, all adversarial clients were detected, while false-positive cases remained, with mean numbers per round of 7, 6, and 7, respectively. These false positives indicate that some honest LCVs can be mistakenly replaced by the estimated reference. We have not quantified the resulting bias in the final contribution scores of honest clients. The detection statistics were identical for the evaluated IID and Non-IID runs, but this does not establish general insensitivity to data heterogeneity.

We also analyze sensitivity to defense/attack hyperparameters in Table~\ref{tab:adversary-defense-sensitivity}: CIF is relatively stable across $\tau$ but more sensitive to $\lambda$, while $v_\text{threshold}$ has a smaller and data-dependent effect. For attack hyperparameters, stronger LCV inflation ($\alpha_a>0.5$) increases CIF under coordinated attacks, while larger PMF magnitudes ($\sigma_a$) increase CIF mainly under IID settings.

\begin{figure}[t]
\centering

% ============================================================
% Top row: Table and Figure
% ============================================================

% Left: Table
\begin{minipage}[c]{0.44\linewidth}
    \centering
    \captionof{table}{Spearman's correlation with the reference
    Monte-Carlo-approximated Shapley value under 25-client settings.}
    \label{tab:n25-fidelity}
    \small
    \setlength{\tabcolsep}{1.2pt}
    \renewcommand{\arraystretch}{0.88}
    \begin{tabular}{llcccc}
        \toprule
        & & \multicolumn{2}{c}{\textbf{Topology}} &
          \multicolumn{2}{c}{\textbf{Data distr.}} \\
        \cmidrule(lr){3-4} \cmidrule(lr){5-6}
        & & cluster & grid & \shortstack{noisy-\\images} & sizes \\
        \midrule

        \multirow{5}{*}{\rotatebox[origin=c]{90}{{\footnotesize TRIP-Shapley}}}
        & vanilla & $0.249$ & $0.136$ & $0.345$ & $0.178$ \\
        & MC      & $0.255$ & $-0.036$ & $0.428$ & $0.108$ \\
        & LOO     & $0.311$ & $0.129$ & $-0.404$ & $-0.168$ \\
        & 8-bit   & $0.298$ & $0.125$ & $0.352$ & $0.152$ \\
        & 4-bit   & \underline{$0.324$} & $0.145$ & $0.311$ & $0.165$ \\

        \midrule

        \multirow{2}{*}{\rotatebox[origin=c]{90}{{\footnotesize DICE}}}
        & 1-hop & $-0.069$ & $0.331$ & $0.475$ & $0.317$ \\
        & 3-hop & $0.065$ & \underline{$0.370$}
               & \underline{$0.591$} & \underline{$0.477$} \\

        \bottomrule
    \end{tabular}
\end{minipage}%
\hfill
% Right: Figure
\begin{minipage}[c]{0.50\linewidth}
    \centering
    % Legend above the two plots
    \includegraphics[width=0.8\linewidth]{
        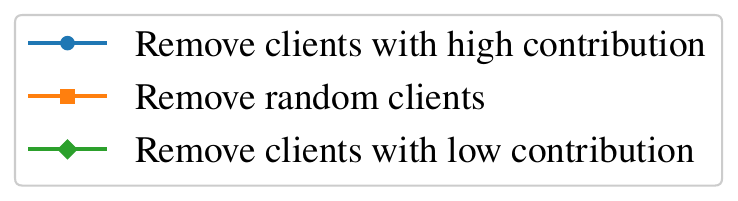
    }
    \vspace{1pt}

    % Two plots
    \begin{subfigure}[t]{0.48\linewidth}
        \centering
        \includegraphics[width=\linewidth]{
            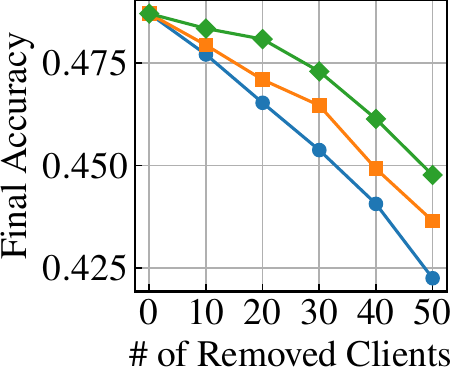
        }
        \caption{CIFAR-10}
    \end{subfigure}%
    \hfill
    \begin{subfigure}[t]{0.48\linewidth}
        \centering
        \includegraphics[width=\linewidth]{
            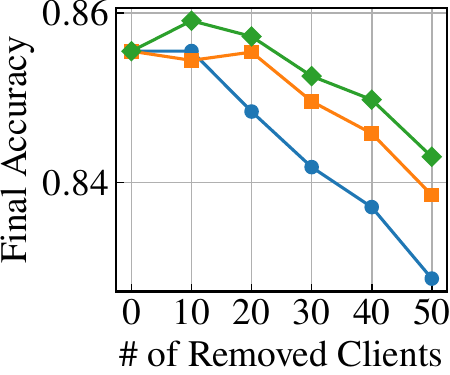
        }
        \caption{FMNIST}
    \end{subfigure}
    \captionof{figure}{Mean final accuracy after removing the $k$ highest-contribution, lowest-contribution, or randomly selected clients.}
\label{fig:client-removal}
\end{minipage}

\vspace{-10pt}
\end{figure}

\begin{figure}[!t]
\centering
\begin{minipage}[t]{0.55\linewidth}
    \centering
    \vspace{0pt}
    
    \captionof{table}{Computation and communication costs for vanilla P2PFL, DICE, and TRIP-Shapley variants under a $d$-regular graph.}
    \scriptsize
    \setlength{\tabcolsep}{1pt}
    \renewcommand{\arraystretch}{0.9}
        \begin{tabular}{@{}llcc@{}}
            \toprule
            \multicolumn{2}{c}{Method} & Comp. & Comm. \\
            \midrule
            \multicolumn{2}{l}{Vanilla P2PFL} & $T_t$ & $dM$ \\
            DICE & ($h$-hop)
                & \makecell{$T_t + $\\$(2 \sum_{i=3}^{h}d^i +3)T_e$}
                & $(d + \sum_{i=1}^{h}d^i)M$ \\
            \midrule
            \multirow{3}{*}{TRIP}
            & Vanilla & $T_t+(d+2^{d+1})T_e$ & $2dM$ \\
            & MC      & $T_t+(d+l)T_e$      & $2dM$ \\
            & LOO     & $T_t+(2d+1)T_e$     & $2dM$ \\
            \bottomrule
        \end{tabular}
        \vspace{1pt}
        \parbox{0.95\linewidth}{\footnotesize
        $T_t$, $T_e$: training/evaluation MAC costs;
        $d$: neighbor count; $M$: model size;
        $l$: number of sampled coalitions in MC.}
    \label{tab:costs}
\end{minipage}%
\hfill
\begin{minipage}[t]{0.44\linewidth}
    \centering
    \vspace{0pt}
    \includegraphics[width=\linewidth]{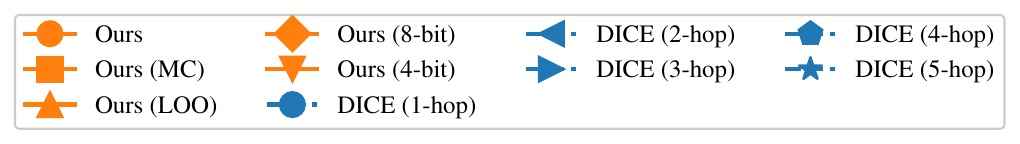}
    \vspace{-4pt}
    \begin{subfigure}[t]{0.48\linewidth}
        \centering
        \vspace{0pt}
        \includegraphics[width=\linewidth]{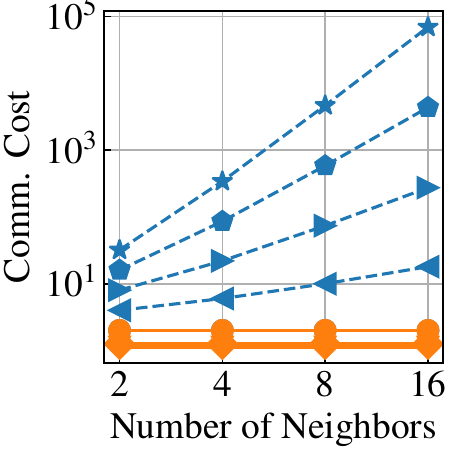}
        \caption{Comm. costs}
    \end{subfigure}
    \hfill
    \begin{subfigure}[t]{0.48\linewidth}
        \centering
        \vspace{0pt}
        \includegraphics[width=\linewidth]{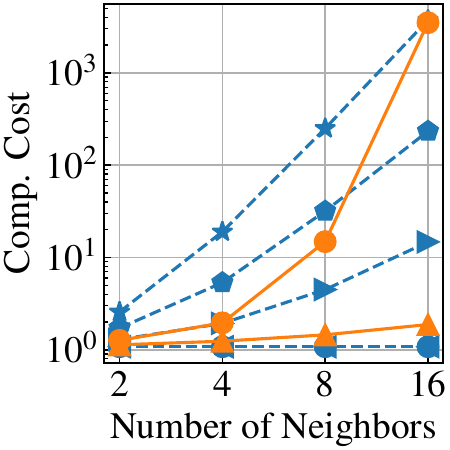}
        \caption{Comp. costs}
    \end{subfigure}
    \captionof{figure}{Cost scaling with the number of neighbors with regular graphs.}
    \label{fig:scale}
\end{minipage}

\vspace{-8pt}
\end{figure}

\subsection{Scalability to Larger Settings}

Finally, we assess the scalability of TRIP-Shapley to larger P2PFL scenarios in terms of both reference alignment and cost. In terms of reference alignment, we have evaluated Spearman's correlation of TRIP-Shapley against Monte-Carlo-approximated Shapley value as a reference, as the exact Shapley value is infeasible with 25-client settings. Additionally, we perform a client removal experiment with 100 clients, a widely used approach in prior studies \citep{ghorbani_data_2019, wang_fast_2024}. We also analyze how computational and communication costs scale with the number of neighbors.

\subsubsection{Rank Alignment with Approximated Reference}

We measure Spearman's correlation with a Monte-Carlo-approximated Shapley value under 25 clients. The reference uses four randomly sampled permutations per setting, corresponding to up to 100 coalition evaluations in total; the random seed varies across settings. This budget does not establish convergence of the reference estimates, so we treat these results as exploratory comparisons against a limited-sample reference. We test (i) 5 fully-connected clusters with central nodes connected in a ring, (ii) a $5\times5$ grid, and (iii) varying noise intensities and data sizes under a 4-regular graph. Table~\ref{tab:n25-fidelity} shows that rank alignment depends on topology and contribution heterogeneity. TRIP variants achieve the highest correlations in the cluster topology, while 3-hop DICE performs best for the grid and both data-heterogeneity settings. TRIP-Shapley's approximation effect is setting-dependent: MC and LOO improve reference alignment in some settings but substantially degrade it in others. The observed comparisons are setting-dependent and should be interpreted subject to the uncertainty of the limited-sample reference.

\subsubsection{Client Removal Experiment with 100 Clients}

In this experiment, we run P2PFL with $k$ clients removed, where the removed clients are selected based on the following three criteria: (i) clients with the highest contributions according to TRIP-Shapley, (ii) clients with the lowest contributions, and (iii) clients selected at random. Selected clients are replaced by dummy clients that perform no local training but continue model exchange, preserving the communication topology. Figure~\ref{fig:client-removal} shows the results. For most values of $k$, the final model accuracy is lowest when high-contribution clients are removed and highest when low-contribution clients are removed, supporting the usefulness of TRIP-Shapley for contribution attribution.

\subsubsection{Computational and Communication Cost Analysis}

Finally, we analyze how computational and communication costs scale with system size. Table~\ref{tab:costs} summarizes the analytical costs. Regarding communication, TRIP-Shapley requires twice that of vanilla P2PFL, and scales linearly with neighbor size $d$. Regarding computation, vanilla TRIP-Shapley requires $2^{d+1}+d$ model evaluations per client per round, while LOO and MC variants require $2d+1$ and $d+l$, respectively, where $l$ is the sampled evaluation budget, enabling a tunable cost--reference-correlation trade-off. Figure~\ref{fig:scale} reports costs relative to vanilla P2PFL for 100 clients with varying $d$. Coalition evaluations can be performed sequentially, so all coalition models need not be retained simultaneously. However, peak memory was not measured and remains implementation-dependent; LOO and MC reduce evaluation count, not necessarily memory usage. Thus, evaluation counts characterize computational scaling rather than practical runtime or memory usage.

\section{Limitations}

\textbf{Reliance on a shared auxiliary evaluation dataset.} Our work assumes that all participants have a common evaluation dataset, which is essential for a consistent evaluation criterion across LCVs from different clients. Indeed, our experiments with 8-client CIFAR-10 settings in Section~\ref{sec:exp-fidelity} show that assigning each client a different test dataset following its training-data distribution degrades the reference correlation from $0.524$ to $0.500$ in the \texttt{noisy-labels} setting and from $0.810$ to $0.071$ in the \texttt{non-iid} setting. Reducing the shared evaluation set from 500 to 100 examples in the 8-client CIFAR-10 setting also degrades the reference correlation, most notably in the \texttt{cluster} ($0.810 \to 0.405$) and \texttt{line} ($0.714 \to 0.190$) settings. Hence, addressing settings without a shared dataset, as well as alternative utility functions or distribution shifts in the shared test dataset, is left to future work.

\textbf{Necessity of an honest coordinator.} While our work assumes fully decentralized training, it relies on a trusted coordinator for contribution attribution. Without such a coordinator, contribution attribution would require Byzantine-tolerant consensus among participants, making communication and synchronization costs impractical at scale. Conversely, an untrusted coordinator could manipulate collected LCVs and produce incorrect attribution. This reliance limits deployment in open or cross-organizational settings where participants lack a common trusted party or are unwilling to entrust it with attribution information. 

\textbf{Local Honest-Majority Assumption.} Our LCV outlier detection relies on the local honest-majority assumption. When violated, a client whose neighbors are predominantly adversarial may receive a majority of corrupted LCVs, making it difficult to distinguish corrupted LCVs from legitimate ones. This may cause the mitigation to fail and constitutes a limitation of our approach.

\textbf{Cross-Recipient LCV Consistency Assumption in LCV Outlier Detection.} Although our LCV outlier detection relies on the cross-recipient LCV consistency assumption, as mentioned in Section~\ref{sec:proposal-adversarial}, honest LCVs for the same client can vary across recipients despite using a common evaluation dataset. In our 25-client all-honest experiments, the mean cross-recipient standard deviation was 0.0163 (95th percentile: 0.0288) under IID data and 0.0779 (95th percentile: 0.1398) under Non-IID data. Thus, recipient-dependent variation can become substantial, especially under heterogeneous data, potentially leading to false outlier detection and biased LCV correction. More robust detection and correction methods that account for recipient, neighborhood, or aggregation weights are left for future work.

\textbf{False Positive Attack Detection Can Distort Honest Clients' Scores.} Relatedly, replacing falsely flagged LCVs with a recipient-independent reference vector $v$ can ignore inherent recipient-dependent differences and thereby introduce bias into the final contribution scores of honest clients. In a preliminary comparison on five clients in the 25-client setting, this bias was small under IID data but more pronounced for some clients under Non-IID data; a systematic quantification across all clients is left for future work. Similarly, pre-training-model filtering can replace honest models with naturally lower accuracy, particularly under Non-IID data; notably, $62.1\%$ of pre-training models were flagged even without attacks in our Non-IID setting at $v_{\mathrm{threshold}}=0$, while the false-replacement rate has not been quantified.

\section{Conclusion} \label{sec:conclusion}

In this study, we proposed TRIP-Shapley, a propagation-aware, Shapley-inspired surrogate for contribution attribution in P2PFL. The experiments demonstrate favorable communication-cost trade-offs, while rank alignment with reference Shapley values depends on the dataset, topology, and approximation variant. The 25-client results remain exploratory because of the limited reference-sampling budget. The defenses reduce contribution inflation under the evaluated attacks, but their impact on honest clients and robustness to adaptive evasion require further validation.

\section*{Acknowledgments}
This work is supported in part by JST CREST JPMJCR21D2.

\bibliography{main}

\end{document}